\documentclass[runningheads]{llncs}
\usepackage{multirow}
\usepackage{graphicx}
\usepackage{epstopdf}
\usepackage{booktabs}
\usepackage{bbding}
\begin{document}
\title{Distinguishing Individual Red Pandas \\from Their Faces}
%
%
\author{Qi He\inst{1} \and
Qijun Zhao\inst{1,a} \and
Ning Liu\inst{1} \and
Peng Chen\inst{2}\and
Zhihe Zhang\inst{2}\and
Rong Hou\inst{2,a} 
}

\authorrunning{Q. He et al.}
%
\institute{College of Computer Science, Sichuan University, Chengdu, Sichuan 610065, China \and
Chengdu Research Base of Giant Panda Breeding, Sichuan Key Laboratory of Conservation Biology for Endangered Wildlife, Chengdu, Sichuan 610086, China
}
\maketitle              

\begin{abstract}
Individual identification is essential to animal behavior and ecology research and is of significant importance for protecting endangered species. Red pandas, among the world's rarest animals, are currently identified mainly by visual inspection and microelectronic chips, which are costly and inefficient. Motivated by recent advancement in computer-vision-based animal identification, in this paper, we propose an automatic framework for identifying individual red pandas based on their face images. We implement the framework by exploring well-established deep learning models with necessary adaptation for effectively dealing with red panda images. Based on a database of red panda images constructed by ourselves, we evaluate the effectiveness of the proposed automatic individual red panda identification method. The evaluation results show the promising potential of automatically recognizing individual red pandas from their faces. We are going to release our database and model in the public domain to promote the research on automatic animal identification and particularly on the technique for protecting red pandas.

\end{abstract}
\footnotetext[0]{\inst{a}Co-corresponding authors:~R.~Hou~(405536517@qq.com),~Q.~Zhao~(qjzhao@scu.edu.cn)}

%
\section{Introduction}
\begin{figure}
\includegraphics[width=\textwidth]{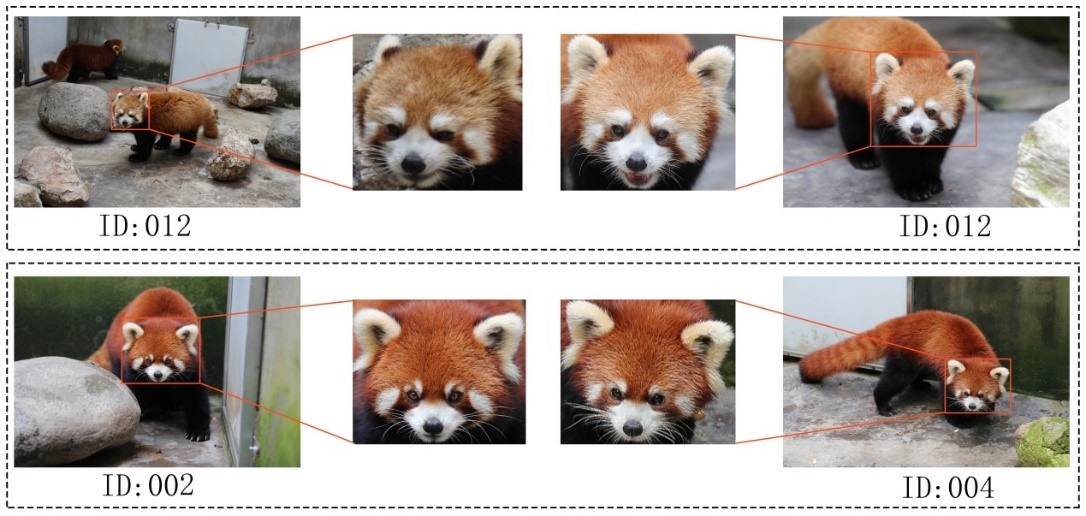}
\caption{Examples of red panda images of the same (top row) and different individuals (bottom row).} \label{fig1}
\end{figure}

\emph{Ailurus fulgens}, also known as lesser panda and red panda (see Fig.~\ref{fig1})), is endemic to the Himalayan - hengduan mountains. They are mainly distributed in China, Nepal, India, Bhutan and Myanmar. It is estimated that there are only 16,000 to 20,000 red pandas in the world while 6400 to 7600 of them are in China. According to~\cite{ref_1}, over the past 50 years, the number of wild red pandas has decreased by about 40\% in China due to habitat loss, human activities and hunting. The population of wild red pandas in the world is also decreasing year by year~\cite{ref_2}. Existing red pandas were listed as endangered species by IUCN in 2000, and they were classified as a Category II species under the Wild Animal Protection Law in China. In order to protect red pandas, it is important to maintain precise and up-to-date information of the population and distribution of red pandas, which requires the technique of distinguishing individual red pandas.

Traditionally, people have to spend many months to collect the excreta and biology samples of wild red pandas, based on which different red pandas are identified. This method is obviously labour-intensive and has long identification cycles. As more and more surveillance cameras are deployed for monitoring animals, people manually identify individual red pandas on the camera-trap images based on their appearance. For captive breeding red pandas, people distinguish individual red pandas either via visual inspection or by scanning the microelectronic microchips implanted in the body of red pandas. However, manual identification and visual inspection are tedious and error-prone, especially for people without training. Identification using microchips is generally more accurate, but as an invasive approach, it is hurtful to red pandas and unfriendly to operate.

With the rapid development of computer vision technology in the past decade, some researchers attempt to automatically identify individuals of specific species based on images of the animals~\cite{ref_3,ref_4,ref_5,ref_6,ref_7,ref_8,ref_9,ref_10,ref_11,ref_12}. Following the pipeline of typical pattern recognition systems, they extract discriminative features from certain body parts of the animals, compute the similarity scores between images of the animals based on the extracted features, and finally determine the identities of the individual animals in the images according to their similarity scores with the reference images. Table~\ref{tab1} summarizes existing research on image-based automatic individual identification of animals. Compared with traditional methods, these automatic methods are not only more friendly but also more efficient. Some of them achieve very promising identification accuracy, which demonstrates the potential of image-based automatic identification of individual animals.

Motivated by these studies, we aim to investigate in this paper the feasibility of automatically distinguishing individual red pandas based on their face images. To this end, we construct an image database of red pandas with labeled identities, which is going to be released for research purpose. By exploring latest deep learning techniques, we develop methods for automatically detecting red panda faces, locating their eyes and noses, and extracting features from the aligned red panda faces. We finally build a framework for automatic identification of individual red pandas, and evaluate its effectiveness on the constructed database. In the rest of this paper, we first review related work on automatic individual identification of animals in Section 2, then introduce in detail our framework for red panda identification in Section 3, followed by evaluation results in Section 4, and finally conclude the paper in Section 5.

\section{Related Work}
As summarized in Table~\ref{tab1}, automatic individual identification methods have been studied for a number of species, including African penguins~\cite{ref_3}, northeast tigers~\cite{ref_4}, cattle~\cite{ref_5}, lemurs~\cite{ref_6}, dairy cows~\cite{ref_7}, great white sharks~\cite{ref_8}, pandas~\cite{ref_9}, primates~\cite{ref_10}, pigs~\cite{ref_11}, and ringed seals~\cite{ref_12}. Different species usually have largely different appearance; however, different individual animals of the same species may differ quite slightly in their appearance, and can be distinguished only by fine-grained detail. Almost all of the related studies are based on specific body parts of an animal to determine its identity. For those species that have salient characteristics in their appearance (e.g., the spots on the breast of penguins~\cite{ref_3}, and the rings on the body of ringed seals~\cite{ref_12}), individual identification can be done by extracting and comparing their salient features. For those species that have subtle appearance differences between different individuals, such as pigs~\cite{ref_11}, lemurs~\cite{ref_6}, and pandas~\cite{ref_9}, the most common solution to individual identification is to focus on the body parts with relatively rich textures and extract discriminative features from the parts. 
%
\begin{table}
\caption{A summary of existing image-based individual identification of animals.}\label{tab1}
\resizebox{\textwidth}{!}{%
\begin{tabular}{c|c|c|c|c|c|c}
\toprule[2pt]
\multirow{2}{*}{\textbf{Species}} & \multirow{2}{*}{\textbf{Discriminative Body Parts}} & \multicolumn{2}{c|}{\textbf{Database}} & \multicolumn{2}{c|}{\textbf{Features}} & \multirow{2}{*}{\textbf{Accuracy}} \\ \cline{3-6}
  &  & \#subjects          & \#images         & Hand-Crafted         & Learned         &   \\ \midrule[1pt]
African penguins~\cite{ref_3}          & Spots on Breast                                     & \multicolumn{2}{c|}{N/A}               & \Checkmark &            & 75.5\% \\  \hline
Northeast tigers~\cite{ref_4}          & Texture on Body                & 103                 & 10300            &                      &\Checkmark               & N/A   \\ \hline
Cattle~\cite{ref_5}                   & Muzzle Print                & \multicolumn{2}{c|}{N/A}               & \Checkmark                  &                 & N/A      \\ \hline
Lemurs~\cite{ref_6}                   & Face                        & 80                  & 462              & \Checkmark            &          & 98.7\%$\pm$1.81\%  \\ \hline
Dairy cows~\cite{ref_7}               & Tailhead                    & 10                  & 1965             & \Checkmark              &                 & 99.7\%     \\ \hline
Great white sharks~\cite{ref_8}        & Fin Shape                  & N/A                 & 240              & \Checkmark                   &                 & N/A   \\ \hline
Pandas~\cite{ref_9}                   & Face                        & 18                  & 131              &                      & \Checkmark              & 58.82\% \\\hline
Primates~\cite{ref_10}                & Face                        & 280                 & 11637            &                      & \Checkmark               & N/A   \\ \hline
Pigs~\cite{ref_11}                     & Face                       & 10                  & 1553             &                      & \Checkmark               & 96.7\% \\ \hline
Ringed seals~\cite{ref_12}            & Rings on Body               & 131                 & 591              & \Checkmark                   &                 & N{/}A  \\ 
\bottomrule[2pt]
\end{tabular}%
}
\end{table}

Red pandas obviously belong to those species that have subtle appearance differences between different individuals. Fortunately, their faces have relatively salient textures. According to Table~\ref{tab1}, most methods for the species that do not have salient appearance differences are based on learned features. With learning based models, researchers do not have to manually find out the exact parts that are helpful to identification. Inspired by these works, we build a deep neural network model for identifying individual red pandas based on their face images. Compared with existing animal identification methods, ours is fully automatic. Almost all existing methods are based on pre-cropped pictures of specific body parts, such as the tailhead images of dairy cows~\cite{ref_7} and face images of pig~\cite{ref_11}. In contrast, our method takes the image of a red panda as input and automatically detect its face, extracts features and matches the features to the ones enrolled in the gallery to determine its identity. In addition, to the best of our knowledge, the research in this paper is the first attempt to image-based automatic individual identification of red pandas. 

\section{Method}
\subsection{Overview}
The proposed automatic individual red panda recognition framework mainly includes three modules: face detection, face alignment and identification. See Fig.2. Given an image of red panda, the first step is to detect whether there is a red panda face in the image or not. If there is a red panda face, its eyes are located, according to which the red panda face image is aligned such that the line connecting the two eyes is horizontal. Finally, features are extracted from the cropped and aligned face image and compared to obtain the red panda’s identity. In the remainder of this section, we introduce the detail of each module.
\begin{figure}
\centering
\includegraphics[width=10cm,height=5cm]{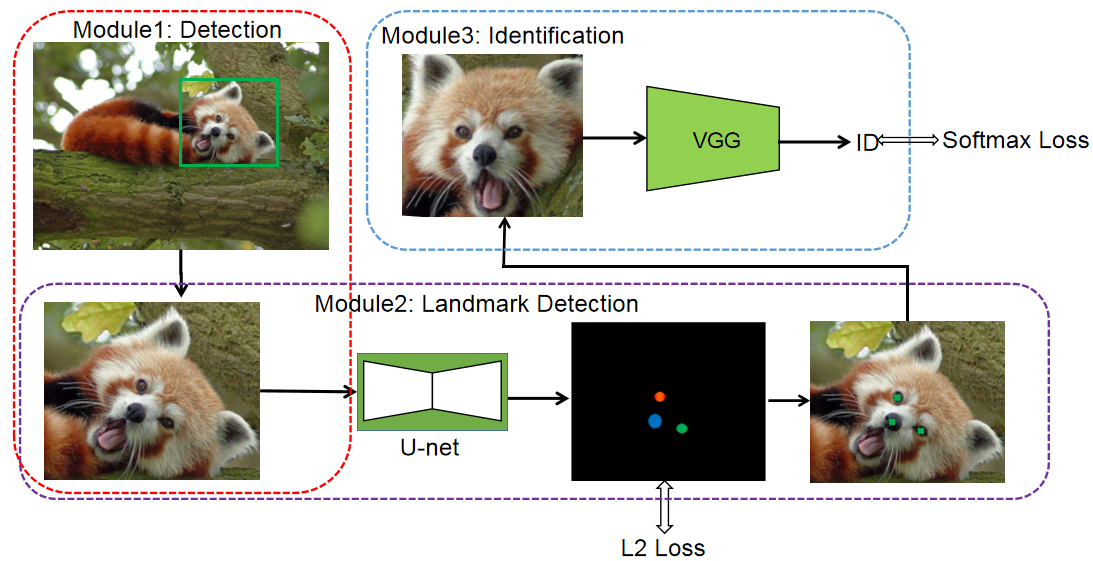}
\caption{The flowchart of proposed method.} \label{fig2}
\end{figure}
\subsection{Red Panda Face Detection}
In this paper, we use the well-known object detection model YOLOv2~\cite{ref_19} to detect red panda faces because of its balance between efficiency and accuracy. We fine-tune the pre-trained YOLOv2 model in~\cite{ref_20} by using our collected red panda images. Using the fine-tuned detection model, the bounding boxes of the red panda faces in the images are obtained. If more than one red panda faces are detected in one image, only the largest one is identified. Red panda face images are then cropped according to the bounding boxes. In these images, the red panda faces could be rotated (see Fig.~\ref{fig2}), and the resulting face images would contain much background that may distract the identification process. Therefore, in the next module, we align and further crop the red panda face images to minimize the influence of posture on the identification of the red pandas.
\subsection{Red Panda Face Alignment}
Like in human face recognition, we align red panda face images based on the centers of the two eyes such that the line connecting the eye centers is horizontal. For this sake, we have first to locate the eye centers on red panda faces. Looking closely at the red panda faces, it is easy to notice that the eyes and nose of red pandas have obvious distinct characteristics compared with other regions on the face; that is, they appear as round and black spots. Inspired by this observation, we locate the centers of eyes and nose on red panda face by segmenting the eye and nose regions rather than directly estimating their coordinates.

Specifically, given a red panda face image, we employ U-Net~\cite{ref_16} to segment two eye regions and one nose region. The output of this U-Net is a three-channel image that has the same size as the input image, and each of its three channels represents one of the three regions (see Fig.~\ref{fig2}). Take one of the eye regions as an example. Its corresponding channel is a binary mask, on which the values of pixels in the eye region equal to 255 and the values of the rest pixels are set to 0. The eye region is defined as a circular region centered at the eye center and of a pre-specified radius. The nose region is defined similarly. The radius of eye regions and nose region is set as 7 and 13 pixels, respectively.

Once the eye regions are segmented, we compute their centroids as the eye centers. Based on the eye centers, we rotate the red panda face image such that the two eye centers are on the same horizontal line. Afterwards, we further crop the face image in the following way. Let the distance between the two eye centers be denoted as d. Then, the distances between the eye centers to the top and the bottom of the cropped face image are a*d and b*d, respectively, and the left and right margins with respect to the left and right eye centers are both c*d. Here, a, b, and c are chosen by experience. To crop the red panda face images, the ratio parameters are set as a=1.3, b=1.7 and c=1.2 in this paper. This way, as can be seen from Fig.~\ref{fig2}, the finally obtained red panda face images are not only free from in-plane rotation, but also less affected by the background in the original red panda images.

Note that the nose center is not utilized when cropping the red panda face images. But we include the nose region during segmentation. This is because nose regions have similar appearance as eye regions and they, if not segmented together with the eye regions, could lead to false detection of eye regions. Besides, they can effectively constrain eye regions from irregular positions because of the relative spatial distribution of eyes and nose on red panda faces.
\subsection{Red Panda Identification}
Given an aligned and cropped red panda face image, we use the VGG-16 network~\cite{ref_17} to extract features for identification. VGG-16 is widely used in human face recognition. Here, we take the VGG-16 model in~\cite{ref_21} as a pre-trained model, and use our collected red panda face images to fine-tune it. The obtained VGG-16 model is applied to extract features from the input red panda face images, and the similarity between two red panda face images is calculated based on the cosine distance between their features.

To determine the identity of an input probe red panda image, its facial feature is first extracted as mentioned above, and then compared with the features of all the red pandas enrolled in the gallery. Its identity is finally determined as the one that has the highest similarity with it. Note that if a similarity threshold is specified and the highest similarity is below the threshold, then the probe red panda image is not from any of the red pandas in the gallery; in other words, it is an unknown new individual red panda.

\section{Experiments}
\subsection{Database}
In order to evaluate the effectiveness of the proposed method, we construct a database of red panda images by ourselves because no such database is available in the public domain. All our data are collected at the Chengdu Research Base of Giant Panda Breeding from three sources: (i) high resolution pictures taken by a professional photographer, (ii) images extracted from videos (one image every ten frames), and (iii) lower resolution pictures taken with mobile phones. Totally, 51 individual red pandas are imaged, whose identity information is obtained by scanning the microchips implanted in their bodies. The total number of acquired red panda images is 7,091. Note that in this paper we require that both eyes of the red pandas should be visible in the images. For each image, we manually mark the bounding box of the red panda face together with three landmarks in it (i.e., nose center, left and right eye centers). See Fig.~\ref{fig1} for example images of individual red pandas in our database.

It is worthy mentioning that the images extracted from the same video might be highly correlated. Consequently, using such images for training, the generalization ability of the obtained model would be poor, while using them for testing, the resulted recognition accuracy could be misleadingly high. Being aware of this problem, we use SSIM~\cite{ref_18} to measure the image-level similarity among the video images of each individual. Starting from a randomly chosen image of an individual, we establish the image set of the individual by gradually adding other images of it if the similarity between the images and the already retained images is smaller than a pre-specified threshold. After processing the images of all the individuals, we finally get a database of 2,877 images of 51 red pandas.

In the following experiments, we randomly choose the images of 34 individual red pandas as training data, while the images of the rest 17 individuals are used for test. In the identification experiments, the gallery consists of all the images of the 34 training individuals and 50\% of the images of the 17 test individuals, and the probe consists of the other 50\% of the images of the 17 test individuals.
\subsection{Identification Accuracy by Different Features}
\begin{figure}
\centering
\includegraphics[width=10cm,height=7cm]{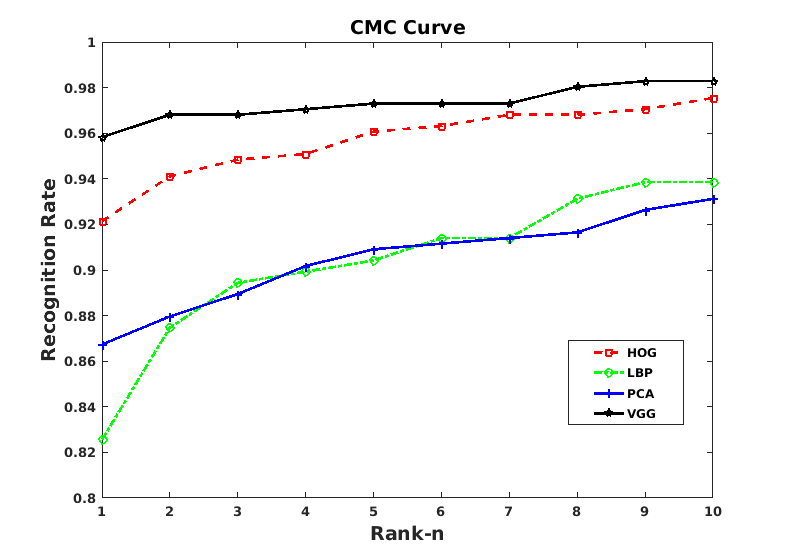}
\caption{Cumulative Match Characteristic (CMC) curves by different feature extraction algorithms.} \label{fig3}
\end{figure}

In this experiment, we compare the identification accuracy of our learned features (denoted by VGG) with that of some representative hand-crafted features, including local binary patterns (LBP)~\cite{ref_14}, histograms of orientation of gradients (HOG)~\cite{ref_15}, and the feature extracted by applying principal component analysis to the red panda face images (PCA)~\cite{ref_13}. These hand-crafted features have been explored by other researchers for identifying individuals of other species. Fig.~\ref{fig3} plots the obtained Cumulative Match Characteristic (CMC) curves. Obviously, our learned features achieve the highest accuracy among the four different feature representations. In all the following experiments, we use the learned VGG features.

\subsection{Performance before and after Discarding Correlated Images}
\begin{figure}
\centering
\includegraphics[width=9cm,height=7cm]{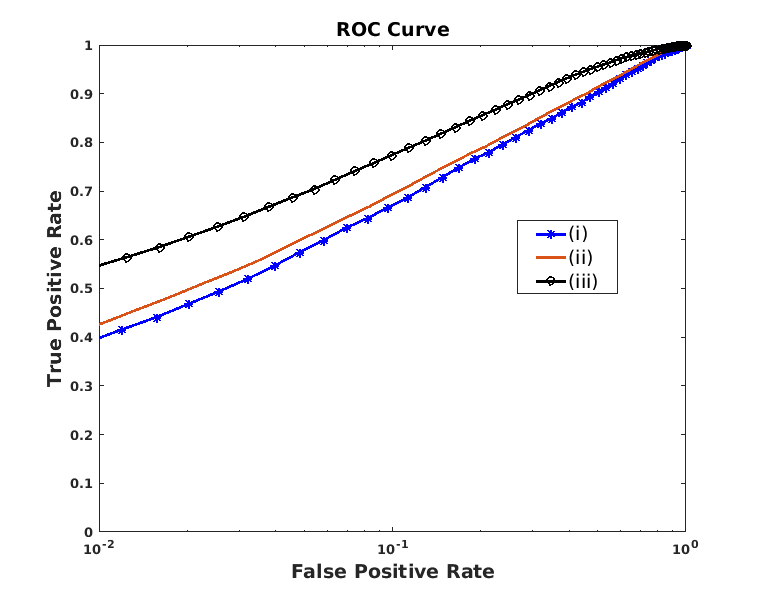}
\caption{The Receiver Operating Characteristic (ROC) curves of the models trained and
evaluated on different data sets. (i) Training with the original image set, and testing on the
cleaned image set, (ii) training and testing both with cleaned image set, (iii) training with the
cleaned image set, while testing on the original image set.} \label{fig4}
\end{figure}

In this experiment, we consider the following three cases: (i) training VGG-16 with the original image set, and testing it on the cleaned image set (i.e., correlated images are discarded), (ii) training and testing VGG-16 both with cleaned image set, (iii) training VGG-16 with the cleaned image set, while testing it on the original image set. To compute the recognition accuracy, we randomly construct 1,000 genuine pairs and 1,000 imposter pairs. The obtained Receiver Operating Characteristic (ROC) curves are shown in Fig.~\ref{fig4}. These results prove the necessity of discarding correlated images in the database to improve the generalization ability of the trained model as well as to make the evaluation more reliable.
\subsection{Impact of Face Alignment}
We first evaluate the precision of our proposed method in localizing the nose and eye centers. We measure the landmark localization precision by the mean squared errors (MSE) between the predicted and the ground truth coordinates of the landmarks on all the images of the 17 test individuals. The results are reported in Table~\ref{tab2}. On average, our proposed method achieves a landmark localization error of about 3 pixels on 224*224 images.
\begin{table}
\caption{Error of landmarks predicted by our model.}\label{tab2}
\resizebox{\textwidth}{!}{%
\begin{tabular}{@{}c|c|c|c|c@{}}
\toprule
\textbf{Landmarks}   & \textbf{Left eye center} & \textbf{Right eye center} & \textbf{Nose center} & \textbf{Average error} \\ \midrule
\textbf{MSE(pixels)} & 3.09                     & 2.98                      & 3.32                 & 3.13                   \\ \bottomrule
\end{tabular}%
}
\end{table}

To assess the impact of alignment on the identification accuracy, we compare the performance of VGG-16 on manually aligned images, automatically aligned images by our method, and original unaligned images. Table~\ref{tab3} presents the three-fold cross validation identification results in terms of rank-1, rank-5 and rank-10 identification rates. As can be seen from the results, alignment is helpful for more accurate identification, and the automatic face alignment method can work well for individual identification of red pandas.
\begin{table}
\caption{The impact of alignment methods on recognition.}\label{tab3}
\resizebox{\textwidth}{!}{%
\begin{tabular}{c|c|c|c}
\toprule
\textbf{Alignment Methods} & \textbf{Rank-1 (\%)} & \textbf{Rank-5 (\%)} & \textbf{Rank-10 (\%)} \\ \midrule
Manually Aligned           & \textbf{93.5$\pm$3.7}    & 97.6$\pm$0.6             & \textbf{98.3$\pm$0.4}     \\
Automatically Aligned      & 93.3$\pm$2.5             & \textbf{97.6$\pm$0.8}    & 98.2$\pm$0.4              \\
Without Alignment          & 91.6$\pm$2.4             & 95.1$\pm$0.7             & 97.2$\pm$0.9              \\ \bottomrule
\end{tabular}%
}
\end{table}

\section{Conclusion}
In this paper, an automatic red panda recognition framework based on facial images is proposed. It effectively utilizes pre-trained deep learning models for processing red panda images, including red panda face detection, alignment and feature extraction, and achieves promising results on a red panda image database constructed by ourselves. However, this paper still has the following limitations. (i) The red panda images are captured in relatively constrained conditions rather than in-the-wild conditions. It is much more challenging but of even more importance to deal with in-the-wild red panda images. (ii) The size of the database is small. The effectiveness of automatic individual identification still needs to be further evaluated on larger scale datasets of more challenging images (e.g., with large variations in pose and illumination). Nevertheless, as the first attempt to automatically identify individual red pandas, this paper presents a promising effort along this direction, and we expect more significant advance as more data become available.
\section*{Acknowledgements}
This research is supported by the National Natural Science Foundation of China (31300306), the Sichuan Science and Technology Program (2018JY0096), Chengdu Research Base of Giant Panda Breeding (CPB2018-02, CPB2018-01), the Chengdu Research Base of Giant Panda Breeding Research Foundation (CPF Research 2014-02, 2014-05), and the Panda International Foundation of the National Forestry Administration, China (CM1422, AD1417). The authors would like to thank Huan Tu for her constructive suggestions on paper writing, thank Zhicong Feng and Chengdong Wang from Sichuan University for their helpful discussion and assistance in data collection, as well as Kongju Wu, Kai Cui and other colleagues from Chengdu Research Base of Giant Panda Breeding for their help with acquiring data of red pandas.

%
%
%

\begin{thebibliography}{21}
\bibitem{ref_1}
Yunfang X.: Genetic Diversity of the Captive Red Panda in China and Paternity. Doctor, Yangzhou University (2015).
\bibitem{ref_2}
The IUCN Red List of Threatened Species, \url{http://dx.doi.org/10.2305/IUCN.UK.2015-4.RLTS.T714A45195924.en}. last accessed 2019/4/18
\bibitem{ref_3}
Burghardt T, Thomas B, Barham P, et al.: Automated visual recognition of individual African penguins. In: Fifth International Penguin Conference, Ushuaia (2004).
\bibitem{ref_4}
Peng Z.: Study on Northeast Tiger Skin Texture Extraction and Recognition Based on BP Network. Doctor, Northeast Forestry University (2008).
\bibitem{ref_5}
Tharwat A , Gaber T , Hassanien A E , et al.: Cattle Identification Using Muzzle Print Images Based on Texture Features Approach. In: Proceedings of the Fifth International Conference on Innovations in Bio-Inspired Computing and Applications IBICA 2014, pp. 217-227. Springer, Cham, (2014).
\bibitem{ref_6}
Crouse D, Jacobs R L, Richardson Z, et al.: LemurFaceID: a face recognition system to facilitate individual identification of lemurs. Bmc Zoology, 2(1), 2 (2017).
\bibitem{ref_7}
Li W, Ji Z, Wang L, et al.: Automatic individual identification of Holstein dairy cows using tailhead images. Computers and Electronics in Agriculture, 142, 622-631 (2017).
\bibitem{ref_8}
Hughes B, Burghardt T.: Automated visual fin identification of individual great white shark. International Journal of Computer Vision, 122(3), 542-557 (2017) 
\bibitem{ref_9}
Houjin, Zheng Bochuan, et al.: Facial recognition of giant pandas based on develop-mental network recognition. ACTA Theriologica Sinicamode, 39(1), 43-51 (2019).
\bibitem{ref_10}
Deb, D., Wiper, S., Gong, S., Shi, Y., Tymoszek, C., Fletcher, A., and Jain, A. K.: Face recognition: Primates in the wild. In: IEEE 9th International Conference on Biometrics Theory, Applications and Systems (BTAS), pp. 1-10. IEEE (2019).
\bibitem{ref_11}
Hansen M F, Smith M L, Smith L N, et al.: Towards on-farm pig face recognition using convolutional neural networks. Computers in Industry, 98, 145-152 (2018).
\bibitem{ref_12}
Chehrsimin T, Eerola T, Koivuniemi M, et al.: Automatic individual identification of Saimaa ringed seals. IET Computer Vision, 12(2), 146-152 (2018).
\bibitem{ref_13}
Jolliffe I T.: Pincipal Component Analysis. Journal of Marketing Research, 25(4), 513 (2002).
\bibitem{ref_14}
Liao S, Zhu X, Lei Z, et al.: Learning multi-scale block local binary patterns for face recognition. In: International Conference on Biometrics, pp. 828-837. Springer, Berlin, Heidelberg (2007).
\bibitem{ref_15}
Dalal N, Triggs B.: Histograms of oriented gradients for human detection. In: Interna-tional Conference on Computer Vision and Pattern Recognition, pp. 886-893. IEEE, Computer Society (2005). 
\bibitem{ref_16}
Ronneberger O, Fischer P, Brox T.: U-net: Convolutional networks for biomedical im-age segmentation. In: International Conference on Medical image computing and computer-assisted intervention, pp. 234-241. Springer, Cham (2015).
\bibitem{ref_17}
Simonyan K, Zisserman A.: Very Deep Convolutional Networks for Large-Scale Image Recognition. In: CoRR abs/1409.1556 (2015).
\bibitem{ref_18}
Wang Z, Bovik A C, Sheikh H R, et al.: Image quality assessment: from error visibility to structural similarity. IEEE transactions on image processing, 13(4), 600-612 (2004).
\bibitem{ref_19}
Joseph Redmon, Ali Farhadi.:YOLO9000: Better, Faster, Stronger. In: 2017 IEEE Con-ference on Computer Vision and Pattern Recognition (CVPR2017), pp.6517-6525. IEEE, Computer Society (2017)
\bibitem{ref_20}
The pretrained YOLOv2 Model, \url{https://pjreddie.com/darknet/yolov2/}. Last accessed 2019/01/30
\bibitem{ref_21}
The pretrained VGG\_FACE.caffemodel, \url{http://www.robots.ox.ac.uk/~vgg/software/vgg\_face/}. Last accessed 2019/01/30
\end{thebibliography}
%

\end{document}